\documentclass[10pt,conference]{IEEEtran}
\IEEEoverridecommandlockouts
\usepackage{cite}
\usepackage{amsmath,amssymb,amsfonts}
\usepackage{graphicx}
\usepackage{textcomp}
\usepackage{xcolor}
\usepackage{microtype}
\usepackage{algorithm}
\usepackage{algpseudocode}
\usepackage{multirow}
\usepackage{booktabs}
\usepackage{cleveref}
\usepackage[section]{placeins}
\def\BibTeX{{\rm B\kern-.05em{\sc i\kern-.025em b}\kern-.08em
    T\kern-.1667em\lower.7ex\hbox{E}\kern-.125emX}}
\begin{document}

\title{Instance-Level Post Hoc Uncertainty Quantification in Object Detection}

\author{\IEEEauthorblockN{Chongzhe Zhang$^{1,2}$, Zifan Zeng$^{1,3}$, Qunli Zhang$^{1}$, Feng Liu$^{1}$, Zheng Hu$^{1}$}
\IEEEauthorblockA{\textit{$^{1}$RAMS Lab, Huawei Heisenberg Research Center, Munich, Germany}\\
\textit{$^{2}$Technische Universität Berlin, Berlin, Germany} \\
\textit{$^{3}$Technische Universität München, Munich, Germany} \\
\{chongzhe.zhang, zifan.zeng, zhangqunli1, feng.liu1, hu.zheng\}@huawei.com}
}

\maketitle
\bstctlcite{IEEEexample:BSTcontrol}

\begin{abstract}
Object detection is a safety-critical component of autonomous driving.
It is essential to quantify the uncertainty in bounding-box predictions for safety assurance.
Post hoc uncertainty quantification without retraining aligns with real-world deployment requirements; therefore, we employ the Laplace approximation.
Because instance-level uncertainty is needed, linearized inference methods that require multiple backpropagations are not time-efficient, and sampling-based methods are not fully post hoc.
We propose Monte-Carlo generalized linearized model (MC-GLM), which provides instance-level and approximately post hoc uncertainty quantification.
The number of samples required in the Monte Carlo step is constant and independent of the number of output instances, so it can be parallelized.
Experiments on the nuScenes dataset with the CenterPoint detector validate the effectiveness of our method, and the resulting uncertainties exhibit good quality.
\end{abstract}


\section{Introduction}
Bounding boxes are the most commonly used form of object position representation in object detection tasks. In safety-critical autonomous driving tasks, accurate quantitative uncertainty of the bounding box is essential for driving safety\cite{feng2019can, timans2024adaptive}. Detection models require a large amount of data in training, and need to be tested and tuned extensively to make the detection model conform to the safety constraints. However, during the training phase, the detection model usually does not contain the ability to quantify the bounding box uncertainty. After the model has been tuned for security, the model cannot be changed. Therefore, in practice, quantifying the uncertainty of a detection model needs to be done without changing the model. In addition, fully post hoc quantification of the bounding box uncertainty is preferred, as it provides a more realistic representation of the uncertainty due to the detection model. For all detection results, we need to individually quantify their uncertainty while ensuring that the process is time-efficient when applied to an autonomous driving system\cite{harakeh2021estimating}.

\subsection{Problem Formulation}
In this paper, we address the uncertainty quantification question to a wide class of models designed for object detection. The aim is to estimate the uncertainty associated with every individual bounding box predicted by the network. We aim to find a solution that satisfies the following requirements:
\begin{itemize}
\item \textbf{Non-intrusive/post hoc}: The method should not affect or alter the original network behaviour, i.e. it should be applicable to a pre-trained model.
\item \textbf{Instance-level}:  Uncertainty estimation at the level of bounding boxes, or finer resolution at the level of features and their correlations.
\item \textbf{Inference speed}: The uncertainty estimation must be usable online for object detection use-cases. 
\item \textbf{Epistemic Uncertainty}: The estimated uncertainty must be the one associated with the lack of knowledge in the trained model.
\end{itemize}

Given the requirement of epistemic and post hoc uncertainty, we are inclined to explore the methods within Bayesian inference via Laplace approximation\cite{yun2023laplace}. Our contribution is in the space of inference methods of estimating the predictive distribution from which the predictive uncertainty is derived. The generalized linearized model (GLM) method guarantees the post hoc nature\cite{glm}, however has impractical compute requirements for bounding box uncertainties. Our method Monte-Carlo generalized linearized model (MC-GLM) is an approximate way to do GLM inference which guarantees limited computing requirements. 

Given an input $x$, a pre-trained network $\theta^*$ and the network function $f_{\theta^*}$, the Laplace approximation method and GLM inference estimates the predictive distribution by 
$$p(z \mid x, D) \approx \mathcal{N} \Big( f(x; \theta^*), J(x)\Sigma_{\theta}J(x)^\top \Big).$$ 
where $J(x) := \left.\frac{\partial f(x;\theta)}{\partial \theta}\right|_{\theta=\theta^*}$ and $\Sigma_{\theta}$ is the covariance of the Laplace approximation normal distribution of the Bayesian posterior. While the covariance $\Sigma_{\theta}$ is computed offline, the Jacobian $J(x)$ must be computed online. This is expensive when $f_{\theta}$ takes values in many dimensions. Our proposed method relies on a Monte-Carlo sampling from the Bayesian posterior in order to replace the Jacobian $J(x)$ by a low-rank and base-changed version $A(x)$ (built from finite-difference directional derivatives along sampled weight perturbations). After column-centering $A(x)$ to obtain $\tilde A(x)$, the predictive covariance is estimated via the (centered) sample covariance as
$$\widehat{\Sigma}_z(x)=\frac{1}{k-1}\,\tilde A(x)\tilde A(x)^\top \approx J(x) \Sigma_{\theta} J(x)^\top.$$
Unlike $J(x)$, which is calculated row-wise by calling back-prop for every scalar value of $f_{\theta}(x)$, the calculation of $A(x)$ is done column-wise by computing approximated directional derivatives along the directions coming from the Monte-Carlo weight samples. This makes the MC-GLM method have limited time requirements, based on the number of Monte-Carlo sampling steps.

% \paragraph{Data and code availability.}
% The implementation used in this work is proprietary and owned by our company, and we are not permitted to publicly release the code or related internal artifacts. To facilitate reproducibility, the paper provides a complete description of the method, including full mathematical derivations and all experimental details.

\section{Related Work}
\subsection{General Uncertainty Quantification}
Accurately quantifying uncertainty in deep neural network predictions has long been a key objective in artificial intelligence. A common approach for turning a deterministic network into a Bayesian neural network is variational inference\cite{gawlikowski2023survey}.
MC Dropout assumes the model parameters follow a Bernoulli variational distribution \cite{gal2015bayesian, gal2016dropout}. Other variational families, including Gaussian distributions, can yield improved performance \cite{graves2011practical, louizos2016structured, sun2017learning, bae2018eigenvalue}. Variational inference has been shown to be applicable to tasks on large-scale datasets\cite{osawa2019practical}.
The Laplace approximation\cite{mackay} can be employed to locally approximate the posterior distribution of the parameters of the trained network in the loss surface \cite{gawlikowski2023survey}. Unlike other techniques, Laplace approximation does not necessitate retraining, but its difficulty lies in computing the Hessian matrix. The computational resource requirement can be reduced by approximating the Hessian matrix using the Fisher information, followed by Kronecker-factored approximation \cite{kfac, george2018fast} or low-rank approximation \cite{lee2020estimating}.
Deep ensembles are also well suited for uncertainty estimation \cite{lakshminarayanan2017simple}. Random initialization, shuffling the training data \cite{lakshminarayanan2017simple}, and hyperparameter random search \cite{wenzel2020hyperparameter} are used to increase model diversity and improve uncertainty estimates when training multiple models.

\subsection{Uncertainty Quantification for Bounding Box}
Recent work has focused on quantifying the uncertainty of bounding-box predictions to improve the safety of detection systems.
Bounding-box uncertainty can be obtained through direct modeling \cite{feng2019can}, and temperature scaling and calibration losses can mitigate miscalibration.
Reference \cite{liu2022autoregressive} proposes to predict the dimensions of the 3D bounding box by autoregression, with quantile boxes employed to quantify the uncertainty in each dimension. 
BayesOD \cite{harakeh2020bayesod} utilizes MC Dropout to compute uncertainty by performing Bayesian inference on bounding boxes generated by greedy clustering. Reference \cite{deepshikha2021monte} points out that, for bounding-box regression with convolutional neural networks, the Monte Carlo DropBlock strategy can better model uncertainty. Deep ensembles have also been applied to quantify uncertainty in bounding boxes \cite{lyu2020probabilistic}.
Reference \cite{timans2024adaptive} proposes to quantify the uncertainty of the bounding box using conformal prediction. The conformal quantile obtained in the classification task is passed to the prediction box regression task and is used to generate the conformal prediction interval for the prediction box.
Laplace approximation has also been applied to uncertainty estimation of the bounding box. 
\cite{gui2022laplace} uses Laplace approximation with diagonal Fisher to obtain the distribution of the network parameters, and the uncertainty estimation of the bounding box was realized on the one-stage detector.

MC Dropout \cite{gal2015bayesian} and deep ensembles \cite{lakshminarayanan2017simple}, while powerful, are unsuitable for strictly post hoc uncertainty because they require training multiple models and produce predictive distributions via sampling.
Non-Bayesian single-pass methods like evidential regression \cite{amini2020deep} also require pre-training or fine-tuning, which shifts predictions away from those of the original MAP model.
Although \cite{gui2022laplace} can provide post hoc bounding-box uncertainty, it suffers from computational inefficiency due to the multiple backpropagation calls required for each bounding box. For this reason, we propose an improvement over GLM that uses a limited number of $M$ forward passes to obtain predictive uncertainties for all boxes at the desired resolution.

\section{Methodology}
\subsection{Bayesian Neural Network}
Consider a neural network $f(\cdot;\theta)$ with parameters $\theta\in\Theta\subseteq\mathbb{R}^{w}$,
mapping an input $x\in X\subseteq\mathbb{R}^{n}$ to an output $z=f(x;\theta)\in Y\subseteq\mathbb{R}^{d}$.
Given a training dataset $D=\{(x_n,y_n)\}_{n=1}^{N}$, Bayesian learning places a prior $p(\theta)$ on
the parameters and yields a posterior $p(\theta\mid D)\propto p(D\mid\theta)\,p(\theta)$.
Let $\theta^*\equiv\theta_{\mathrm{MAP}}$ denote the maximum-a-posteriori (MAP) estimate,
\begin{equation*}
\theta^*=\arg\max_{\theta\in\Theta}\Big(\log p(D\mid\theta)+\log p(\theta)\Big),
\end{equation*}
which in practice is obtained by minimizing the negative log-posterior,
$\mathcal{L}(D;\theta)=-\log p(D\mid\theta)-\log p(\theta)$.
For a test input $x$, the posterior predictive distribution marginalizes parameter uncertainty:
\begin{equation*}
p(z\mid x,D)=\int p(z\mid x,\theta)\,p(\theta\mid D)\,d\theta.
\end{equation*}
Define the Bayesian predictive mean
\begin{equation}
\hat z(x)=\mathbb{E}_{z\sim p(z\mid x,D)}[z]
=\mathbb{E}_{\theta\sim p(\theta\mid D)}\big[f(x;\theta)\big].
\end{equation}
Consider a detector with fixed parameters used at inference time, and suppose we wish to quantify uncertainty in its output without changing the model. Conditioned on a fixed parameter vector $\theta$, the detector is deterministic which always returns $z=f(x;\theta)$. Hence, in our post hoc setting we focus on epistemic uncertainty rather than modeling aleatoric noise. We define the predictive covariance as the epistemic uncertainty:
\begin{equation}
\Sigma_{z}(x)=\mathrm{Cov}_{z\sim p(z\mid x,D)}[z]
=\mathrm{Cov}_{\theta\sim p(\theta\mid D)}\big[f(x;\theta)\big].
\end{equation}

\subsection{Monte Carlo Integration: Uncertainty Estimation by Bayesian Averaging}
A Monte Carlo estimate of the predictive distribution is obtained by first sampling weights
$\theta_i\stackrel{\text{i.i.d.}}{\sim}p(\theta\mid D)$. This produces predictive samples
$z_i(x):=f(x;\theta_i)\in\mathbb{R}^d$. The sample mean estimates the Bayesian prediction:
\begin{equation}
\hat{z}_{MC}(x) = \frac{1}{k} \sum_{i=1}^{k} z_{i}(x) \in \mathbb{R}^d
\end{equation}
The sample covariance estimates the predictive uncertainty:
\begin{equation}
    \hat{\Sigma}_{z,MC}(x) = \frac{1}{k-1} \sum_{i=1}^k \delta_i(x) \delta_i(x)^\top \in \mathbb{R}^{d \times d}
    \label{eq:cov_estimator}
\end{equation}
where $\delta_i(x) = z_{i}(x) - \hat{z}_{MC}(x)$.
One successful method following this principle is MC Dropout \cite{gal2015bayesian, gal2016dropout}, where the posterior is estimated by variational inference with Bernoulli distributions.
A key disadvantage of Monte Carlo sampling is that it is not post hoc: the estimated $\hat{\Sigma}_{z,MC}(x)$ quantifies uncertainty for the Bayesian prediction $\hat{z}_{MC}(x)$, which in general does not equal the original MAP prediction $f(x;\theta^*)$.

\subsection{Posterior Distribution by Laplace Approximation}
Laplace approximation \cite{mackay} replaces the intractable parameter posterior by a local Gaussian
approximation around the MAP estimate $\theta^*$:
\begin{equation}
\label{eq:laplace_posterior}
 p(\theta \mid D)\ \approx\ \mathcal{N}(\theta^*, \Sigma_\theta),
\end{equation}
where the covariance is the inverse Hessian of the negative log-posterior,
$\Sigma_{\theta}:=H^{-1}$ with
$H:=\left.\nabla_{\theta}^{2}\mathcal{L}(D;\theta)\right|_{\theta=\theta^*}$.

In practice, $H$ is approximated by the generalized Gauss--Newton matrix, equivalently the (empirical)
Fisher information matrix. Let $g(x,y;\theta):=\nabla_\theta \mathcal{L}((x,y);\theta)$ denote the
per-sample gradient. Then
\begin{equation}
\label{eq:fisher_def}
\begin{aligned}
I_{\theta}(\theta^*)
&:= \mathbb{E}_{(x,y)\sim D}\!\left[g(x,y;\theta^*)\,g(x,y;\theta^*)^\top\right],\\
\Sigma_{\theta}
&\approx I_{\theta}(\theta^*)^{-1}.
\end{aligned}
\end{equation}

\subsection{Fisher Information Estimation via KFAC}\label{kfac_section}
The Fisher information matrix $I_\theta(\theta^*)\in\mathbb{R}^{w\times w}$ is typically too large to
form explicitly. Kronecker-factored Approximate Curvature (KFAC) \cite{kfac, george2018fast} makes two standard approximations: (i) it ignores
cross-layer blocks (block-diagonal across layers), and (ii) it factorizes each layer block as a
Kronecker product.

Concretely, for layers $\ell=1,\ldots,L$ with $w_\ell$ parameters (and $\sum_{\ell=1}^L w_\ell=w$), KFAC
approximates the Fisher by
\[
I_\theta(\theta^*) \approx \operatorname{blkdiag}\big(I_{1},\ldots,I_{L}\big),
\qquad I_{\ell}\in\mathbb{R}^{w_\ell\times w_\ell}.
\]
For a layer with weight matrix $W_\ell\in\mathbb{R}^{\ell_{\mathrm{out}}\times \ell_{\mathrm{in}}}$, each block
is further approximated as
\[
I_{\ell} \approx Q_{\ell}\otimes H_{\ell},
\qquad
Q_{\ell}\in\mathbb{R}^{\ell_{\mathrm{in}}\times \ell_{\mathrm{in}}},\ 
H_{\ell}\in\mathbb{R}^{\ell_{\mathrm{out}}\times \ell_{\mathrm{out}}}.
\]
Combining both gives the Kronecker-factored, block-diagonal approximation
\[
I_{\theta,\mathrm{KFAC}}
:= \operatorname{blkdiag}\big(Q_{1}\otimes H_{1},\ldots,Q_{L}\otimes H_{L}\big).
\]
Using $(Q\otimes H)^{-1}=Q^{-1}\otimes H^{-1}$, the corresponding KFAC posterior covariance is
\begin{equation}\label{sigma_kfac}
\Sigma_{\theta,\mathrm{KFAC}}
:= I_{\theta,\mathrm{KFAC}}^{-1}
= \operatorname{blkdiag}\!\big(
Q_{1}^{-1}\otimes H_{1}^{-1},\ldots,
Q_{L}^{-1}\otimes H_{L}^{-1}
\big).
\end{equation}

\subsection{Generalized Linear Model for Post Hoc Uncertainty}
A generalized linear model (GLM) is obtained by first-order Taylor expansion of the network around the
MAP estimate $\theta^*$:
\begin{equation}
\label{eq:glm_linearization}
f_{\mathrm{lin}}(x;\theta)=f(x;\theta^*)+J(x)\,(\theta-\theta^*),
\end{equation}
where $J(x):=\left.\frac{\partial f(x;\theta)}{\partial \theta}\right|_{\theta=\theta^*}\in\mathbb{R}^{d\times w}$.

Under the Laplace posterior $\theta\sim\mathcal{N}(\theta^*,\Sigma_\theta)$, the affine map
\eqref{eq:glm_linearization} induces a Gaussian predictive distribution for
$z=f_{\mathrm{lin}}(x;\theta)$ with
\begin{equation}
\hat z_{\mathrm{lin}}(x)=f(x;\theta^*),
\qquad
\Sigma_{z,\mathrm{lin}}(x)=J(x)\,\Sigma_\theta\,J(x)^\top.
\label{glm_unc}
\end{equation}

The key post hoc property is that the predictive mean matches the original model output at $\theta^*$,
while the covariance \eqref{glm_unc} provides an analytic expression for epistemic predictive uncertainty.

\subsection{Computation of GLM Uncertainty with KFAC Posterior}
To evaluate the GLM uncertainty $\Sigma_{z,\mathrm{lin}}(x)$ at inference time, this section requires (i)
a parameter-space covariance $\Sigma_\theta$ (computed offline; in this work, $\Sigma_{\theta,\mathrm{KFAC}}$)
and (ii) the input-dependent Jacobian $J(x)$.

Let $z(x):=f(x;\theta^*)\in\mathbb{R}^d$ denote the vector-valued network output with scalar components
$z_j(x)$, $j=1,\ldots,d$. The Jacobian is defined by stacking parameter gradients,
$J(x):=\big[(\nabla_\theta z_1(x))^\top;\ldots;(\nabla_\theta z_d(x))^\top\big]\in\mathbb{R}^{d\times w}$.
Computing all $d$ rows naively would require $d$ separate backward passes (one per scalar output),
which becomes prohibitive when $d$ is large.

With the KFAC posterior covariance, the GLM predictive covariance is
\begin{equation}\label{glm_unc_kfac}
\Sigma_{z,\mathrm{lin}}(x)
= J(x)\,\Sigma_{\theta,\mathrm{KFAC}}\,J(x)^\top,
\end{equation}
where $\Sigma_{\theta,\mathrm{KFAC}}$ is block-diagonal with Kronecker-factored blocks
\eqref{sigma_kfac}. In practice, the full $d\times d$ matrix is not formed and only marginal variances are retained,
\begin{equation}\label{glmunc}
\Sigma_{z,\mathrm{diag}}(x):=\operatorname{diag}(c_1,\ldots,c_d).
\end{equation}

Using the Kronecker structure of both the per-output gradients and the KFAC covariance blocks, each
variance $c_j$ can be computed in closed form as
\begin{equation*}
c_j
=
\sum_{\ell=1}^{L}
\Big(a_{\ell-1}^{\,j}\,Q_{\ell}^{-1}\,(a_{\ell-1}^{\,j})^\top\Big)\,
\Big(g_{\ell}^{\,j}\,H_{\ell}^{-1}\,(g_{\ell}^{\,j})^\top\Big),
\end{equation*}
where $a_{\ell-1}^{\,j}$ and $g_{\ell}^{\,j}$ denote the forward activations and backpropagated gradients
(at $\theta^*$) for output component $z_j$ in layer $\ell$.

\subsection{MC-GLM: Estimating the GLM Predictive Uncertainty via Monte Carlo Sampling}

The GLM predictive covariance in \eqref{glm_unc} can be written as
\begin{equation}
\label{eq:glm_cov_as_cov}
J(x)\,\Sigma_\theta\,J(x)^\top
= \mathrm{Cov}_{\eta \sim \mathcal{N}(0, \Sigma_\theta)}\!\left[J(x)\eta\right].
\end{equation}
Let $\eta_i \stackrel{\text{i.i.d.}}{\sim} \mathcal{N}(0,\Sigma_\theta)$ and define
$u_i:=J(x)\eta_i\in\mathbb{R}^d$ and $\bar u:=\frac{1}{k}\sum_{i=1}^{k}u_i$. Then a Monte Carlo estimate is
\begin{equation}
\label{eq:mc_cov_u}
J(x)\,\Sigma_\theta\,J(x)^\top
\approx \frac{1}{k-1}\sum_{i=1}^{k}\big(u_i-\bar u\big)\big(u_i-\bar u\big)^\top.
\end{equation}
Although $\mathbb{E}[u_i]=0$, centering is retained to obtain the standard sample covariance and to
mitigate finite-$k$ effects (as well as finite-difference error) that can induce a nonzero sample mean.

Since $J(x)\eta$ is the directional derivative of $f$ with respect to $\theta$ at $\theta^*$, it can be
approximated by finite differences:
\begin{equation}\label{fd_dirder}
\bar D_{\eta}f(x;\theta^*)
:=\frac{f(x;\theta^*+\varepsilon\,\eta)-f(x;\theta^*)}{\varepsilon}.
\end{equation}
Using $k$ samples $\{\eta_i\}_{i=1}^k$, form the matrix of approximate directional derivatives
\begin{equation}\label{A_def}
A(x):=\big[\bar D_{\eta_1}f(x;\theta^*)\ \cdots\ \bar D_{\eta_k}f(x;\theta^*)\big]\in\mathbb{R}^{d\times k}.
\end{equation}
Let $\mathbf{1}_k\in\mathbb{R}^{k}$ denote the all-ones vector and define the column-centered matrix
\begin{equation}
\tilde A(x):=A(x)-\frac{1}{k}A(x)\mathbf{1}_k\mathbf{1}_k^\top.
\end{equation}
The MC-GLM predictive covariance estimator is
\begin{equation}\label{mcglm_est}
\widehat{\Sigma}_{z,\mathrm{MC\text{-}GLM}}(x):=\frac{1}{k-1}\,\tilde A(x)\,\tilde A(x)^\top.
\end{equation}
Algorithm~\ref{mcglm_pcode} summarizes the corresponding online procedure.

To interpret \eqref{mcglm_est}, let $\Sigma_\theta = B B^\top$ and draw $\xi_i\sim\mathcal{N}(0,I_{w})$,
with $\eta_i=B\xi_i$. Stacking $\xi_i$ into $\Xi=[\xi_1,\ldots,\xi_k]\in\mathbb{R}^{w\times k}$ yields
$A(x)\approx J(x)B\Xi$ and thus $\tilde A(x)\approx J(x)B\tilde\Xi$, where
$\tilde\Xi:=\Xi-\frac{1}{k}\Xi\mathbf{1}_k\mathbf{1}_k^\top$. Therefore,
$\widehat{\Sigma}_{z,\mathrm{MC\text{-}GLM}}(x)$ is a low-rank Monte Carlo estimate of
$J(x)\Sigma_\theta J(x)^\top$.

\subsection{Sampling from the KFAC Posterior}\label{sec:kfac_sampling}
In MC-GLM, set $\Sigma_\theta=\Sigma_{\theta,\mathrm{KFAC}}$ from \eqref{sigma_kfac}. To sample
$\eta \sim \mathcal{N}(0,\Sigma_\theta)$, use the reparameterization
$\eta = B\,\xi$, where $\xi\sim\mathcal{N}(0,I_w)$ and $B B^\top = \Sigma_\theta$.
The factor $B=\mathrm{ch}(\Sigma_\theta)$ is not formed explicitly; instead, sampling exploits the
block-diagonal and Kronecker structure.

For each layer $\ell\in\{1,\ldots,L\}$, the corresponding KFAC covariance block is
\begin{equation}
\Sigma_{\theta,\mathrm{KFAC}}^{(\ell)}
= Q_\ell^{-1}\otimes H_\ell^{-1}.
\end{equation}
Let $L_{Q_\ell}=\mathrm{ch}(Q_\ell)$ and $L_{H_\ell}=\mathrm{ch}(H_\ell)$ denote Cholesky factors,
so that $Q_\ell=L_{Q_\ell}L_{Q_\ell}^\top$ and $H_\ell=L_{H_\ell}L_{H_\ell}^\top$. Then
\[
\mathrm{ch}\!\big(\Sigma_{\theta,\mathrm{KFAC}}^{(\ell)}\big)= L_{Q_\ell}^{-T}\otimes L_{H_\ell}^{-T}.
\]
Draw $\xi_\ell\sim\mathcal{N}(0,I_{w_\ell})$ and form the layer-wise sample
\begin{equation}
\label{eq:kfac_layer_sample}
\eta_\ell
= \big(L_{Q_\ell}^{-T}\otimes L_{H_\ell}^{-T}\big)\,\xi_\ell.
\end{equation}
Equivalently, when reshaping the vector $\xi_\ell$ into a matrix of shape
$(\ell_{\mathrm{out}},\ell_{\mathrm{in}})$,
\begin{equation}
\Delta W_\ell
= L_{H_\ell}^{-T}\;\mathrm{reshape}\!\big(\xi_\ell;\ell_{\mathrm{out}},\ell_{\mathrm{in}}\big)\;L_{Q_\ell}^{-1},
\end{equation}
where $\Delta W_\ell$ matches the shape of the weight matrix in layer $\ell$.
Concatenating $\{\eta_\ell\}_{\ell=1}^L$ yields $\eta\sim\mathcal{N}\!\big(0,\Sigma_{\theta,\mathrm{KFAC}}\big)$.

\subsection{Quantifying Bounding Box Uncertainties}
Each 3D bounding box is parameterized by
\[
\beta = [p_x,p_y,p_z,w,l,h,\psi,v_1,v_2]\in\mathbb{R}^9,
\]
where $(p_x,p_y,p_z)$ denotes the box center in world coordinates, $(w,l,h)$ its dimensions,
$\psi$ its yaw angle, and $(v_1,v_2)$ the velocity components.

Given an input $x$, the detector outputs $m=m(x)$ boxes $\{\beta_i(x)\}_{i=1}^{m}$. Stack them into $B(x)\in\mathbb{R}^{m\times 9}$ with $B(x)_{i,:}=\beta_i(x)$ and flatten to obtain $z(x)\in\mathbb{R}^{d}$ with $d=9m$.

The full predictive covariance of $z(x)$ is a $d\times d$ matrix, capturing correlations both across
boxes and across the 9 features. For interpretability and efficiency, MC-GLM retains within-box feature
correlations and ignores cross-box correlations, yielding a per-box covariance tensor
$\{\Sigma_i(x)\}_{i=1}^{m}$ with $\Sigma_i(x)\in\mathbb{R}^{9\times 9}$.

Concretely, reshape the finite-difference matrix $A(x)\in\mathbb{R}^{d\times k}$ from \eqref{A_def} into
$\mathcal{A}(x)\in\mathbb{R}^{m\times 9\times k}$, where $\mathcal{A}(x)_{i,:,:}\in\mathbb{R}^{9\times k}$
collects the $k$ directional derivatives for box $i$. Let
$\tilde{\mathcal{A}}(x)_{i,:,:}=\mathcal{A}(x)_{i,:,:}-\frac{1}{k}\mathcal{A}(x)_{i,:,:}\mathbf{1}_k\mathbf{1}_k^\top$
be column-centered. Then the per-box covariance estimate is
\begin{equation}
\label{eq:per_box_cov}
\widehat{\Sigma}_i(x)=\frac{1}{k-1}\,\tilde{\mathcal{A}}(x)_{i,:,:}\,\tilde{\mathcal{A}}(x)_{i,:,:}^\top\in\mathbb{R}^{9\times 9}.
\end{equation}
For scalar scoring and visualization, summarize each box covariance by
$u_i(x)=\mathrm{tr}(\widehat{\Sigma}_i(x))$, producing a per-box uncertainty vector
$u(x)\in\mathbb{R}^{m}$.

With MC-GLM, the number of network evaluations per input is $k{+}1$ (one for $f(x;\theta^*)$ and $k$
for perturbed weights), independent of $m$.
In contrast, computing per-output marginal variances via the diagonal-GLM approach requires
$d=9m$ separate backward passes (one per scalar output component), making instance-level uncertainty
computationally expensive when $m$ is large.

\begin{algorithm}[tb]
\caption{MC-GLM uncertainty (online): Monte Carlo estimate of
$\Sigma_z(x)\approx J(x)\,\Sigma_{\theta,\mathrm{KFAC}}\,J(x)^\top$}\label{mcglm_pcode}
\begin{algorithmic}[1]
\State $x \gets$ input
\State $\varepsilon \gets \text{float},\ k \gets \text{int}$ \Comment{hyperparameters}
\State $\theta^* \gets$ pre-trained weights \Comment{$\theta^*\equiv\theta_{\mathrm{MAP}}$}
\State $\Sigma_{\theta,\mathrm{KFAC}} \gets$ KFAC posterior covariance (offline)
\State $z_0 \gets f(x;\theta^*)$ \Comment{MAP prediction, size $(d,)$}
\State Initialize $A\in\mathbb{R}^{d\times k}$ \Comment{columns store finite differences}
\For{$i=1$ \textbf{to} $k$}
    \State Sample $\eta_i \sim \mathcal{N}(0,\Sigma_{\theta,\mathrm{KFAC}})$ \Comment{via Sec.~\ref{sec:kfac_sampling}}
    \State $z_i \gets f(x;\theta^* + \varepsilon\,\eta_i)$
    \State $A_{:,i} \gets (z_i - z_0)/\varepsilon$
\EndFor
\State $\tilde A \gets A - \frac{1}{k}A\,\mathbf{1}_k\mathbf{1}_k^\top$ \Comment{column-centering}
\State $\widehat{\Sigma}_z(x) \gets \frac{1}{k-1}\,\tilde A\,\tilde A^\top$
\State \Return $\widehat{\Sigma}_z(x)$
\end{algorithmic}
\end{algorithm}

\section{Evaluations}
The quality of the uncertainty is evaluated by comparison with accuracy. The predictive distribution should in principle produce accurate confidence intervals for containing the ground truth. If the predictive distribution covariance is summarized by a single scalar (e.g. trace, entropy), the uncertainty value should be proportional to the variability of the error. In other words, high error implies high uncertainty (Note that the opposite implication is not true).

\subsection{Metrics}
We use the evaluation metrics proposed by \cite{mukhoti2018evaluating}. The input to the evaluations consists of a validation dataset $D$, which consists of all the prediction instances and their corresponding errors with the ground truths, and uncertainties calculated on the predictions of the pre-trained model. The metrics we use for uncertainty evaluation depend on dividing the predictions (or instances of predictions) into four partitions: $AC$ (accurate \& certain), $IC$  (inaccurate \& certain), $AU$ (accurate \& uncertain) and $IU$ (inaccurate \& uncertain). We also need an error threshold and an uncertainty threshold to define the certain-uncertain and accurate-inaccurate partitions. The uncertainty threshold is taken to be the mean uncertainty over the validation dataset. The error threshold is taken to be the mean error over the validation dataset in case the error is $\ell_2$ loss, or $0.4$ if the error is IoU.

The following metrics measure the quality of uncertainty. 
\begin{enumerate}
\item Probability of accurate given certain; $P(A|C) := \frac{|AC|}{|AC|+|IC|}$
\item Probability of uncertain given inaccurate; $P(U|I) := \frac{|IU|}{|IC|+|IU|}$
\item Accuracy vs Uncertainty; $AvU = \frac{|AC|+|IU|}{|D|}$
\end{enumerate}

The model with a higher value for the above metrics is better. High values for the first two metrics $P(A|C)$ and $P(U|I)$ capture the right properties of the uncertainty-accuracy relationship. A high value of the third metric $AvU$, although desirable, is not necessarily expected from a well-behaved uncertainty model. For example, the $AvU$ can be diminished due to a high density of points in the $AU$ region, something the uncertainty model in principle cannot control. 

As the metric values depend on the uncertainty threshold parameter, it is useful to see how the values vary with this parameter. This gives plots with uncertainty thresholds on the $x$-axis and the corresponding probability values on the $y$-axis. We can measure the difference in quality by observing the gap between the curves corresponding to different models.

\subsection{Experimental Setup}

We select a representative point cloud-based 3D detector CenterPoint\cite{yin2021center} and conduct experiments on the widely used nuScenes dataset\cite{caesar2020nuscenes}.
First, the offline step of computing the KFAC Fisher information is carried out for CenterPoint using the original nuScenes training set, collecting gradients of the training loss at the pre-trained model weights, and computing the sample covariance after the simplification offered by Kronecker factorization of the gradient tensors \cite{kfac}.
We compute the Fisher information matrix for the final layer (i.e. bbox heads) of the models. In the end, we run experiments to understand the effects of this choice by comparing the uncertainty evaluation metrics with Fisher matrices of different depths of the network. For the runtime estimation of predictive covariance using MC-GLM there are two hyperparameters $k$ (the number of MC samples) and $\epsilon$ (for the difference approximation of derivative). We choose $k = 10$ and $\epsilon = 1 \times 10^{-7}$ for all the experiments. 

For the comparison baseline, we use Monte Carlo Integration (MCI) instantiated with MC Dropout \cite{gal2015bayesian, gal2016dropout}, keeping dropout active at test time and performing $k=10$ stochastic forward passes per input. The predictive mean is their average, and the predictive uncertainty is their sample covariance. For the GLM baseline, we use Laplace-GLM inference, where the detector is linearized at the pre-trained weights $\theta^*$ and the predictive covariance is computed as $J(x)\Sigma_{\theta}J(x)^\top$, where $J(x)$ is the output Jacobian and $\Sigma_{\theta}$ is the Laplace posterior covariance estimated offline from the KFAC Fisher matrix. GLM provides instance-level post hoc uncertainty for the original MAP predictions, but requires repeated backpropagation.

\subsection{Experimental Results}

\begin{figure*}[htpb]
	\centering
	\includegraphics[scale=.45]{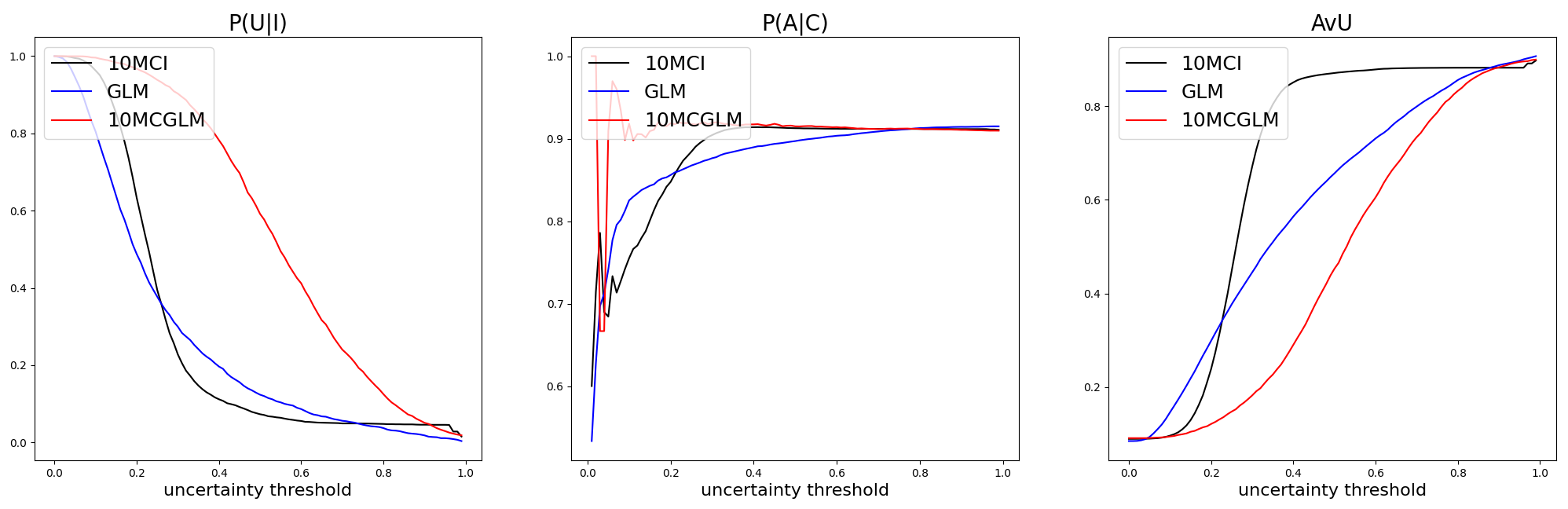}
	\caption{CenterPoint results: plots of $P(U|I)$, $P(A|C)$ and $AvU$ for variable uncertainty thresholds. Each box shows one of these plots for three uncertainty methods (MCI, GLM and MC-GLM). }
	\label{cross model plot}
 \vspace{-5pt}
\end{figure*}

\begin{table}[tb]
\caption{Uncertainty quality comparison across detectors.}
\centering
\begin{tabular}{lccccc}
\toprule
                          &            & MCI     & GLM   & MC-GLM  \\ \hline
\multirow{2}{*}{$AvU$}    & IoU        & 0.87    & 0.64  & 0.62    \\
                          & $\ell_2$   & 0.90    & 0.66  & 0.62    \\ \hline
\multirow{2}{*}{$P(A|C)$} & IoU        & 0.92    & 0.92  & 0.92    \\
                          & $\ell_2$   & 0.95    & 0.93  & 0.95    \\ \hline
\multirow{2}{*}{$P(U|I)$} & IoU        & 0.09    & 0.14  & 0.40    \\
                          & $\ell_2$   & 0.10    & 0.10  & 0.40    \\
\bottomrule
\end{tabular}
\label{tab:uq_results}
\vspace{-10pt}
\end{table}

Tab.~\ref{tab:uq_results} reports uncertainty quality for the point-cloud 3D detector CenterPoint. We compare MCI, GLM and MC-GLM using the metrics $AvU$, $P(A|C)$ and $P(U|I)$, and report results for two base error definitions, IoU and $\ell_2$.
Overall, MC-GLM is a reliable and fast replacement for gradient-based GLM for inference-time predictive uncertainty calculation. In particular, MC-GLM achieves substantially higher $P(U|I)$ than GLM. The Bayesian averaging method MCI has comparable scores in some cases, but is not post hoc, i.e., it does not produce uncertainties corresponding to the original MAP bounding boxes.
In \Cref{cross model plot}, we use the Gamma distribution to fit the uncertainty obtained by each method and replace the original uncertainty value with the quantile to scale all uncertainties to $[0, 1]$. As can be seen from the interval of the curves, MC-GLM can achieve better results than other methods in $P(U|I)$ and $P(A|C)$, which conform to the nature of uncertainty. The lower values for the metric $AvU$ for MC-GLM can be attributed to the high density of samples in the $AU$ partition.

In terms of speed, the MC-GLM run-time is restricted to a preset number of forward passes, $10$ in our case. Compared to this one needs to perform $m \times 9$ backwards passes, where $m$ is the number of detected bounding boxes, to get the same amount of instance-level uncertainty information with GLM. In this experiment, the average number of detected boxes is $m=97.3$. From the results in \Cref{tab:running time}, MC-GLM has a clear advantage over GLM in speed.

\begin{figure}[htbp]
	\centering
	\includegraphics[scale=.35]{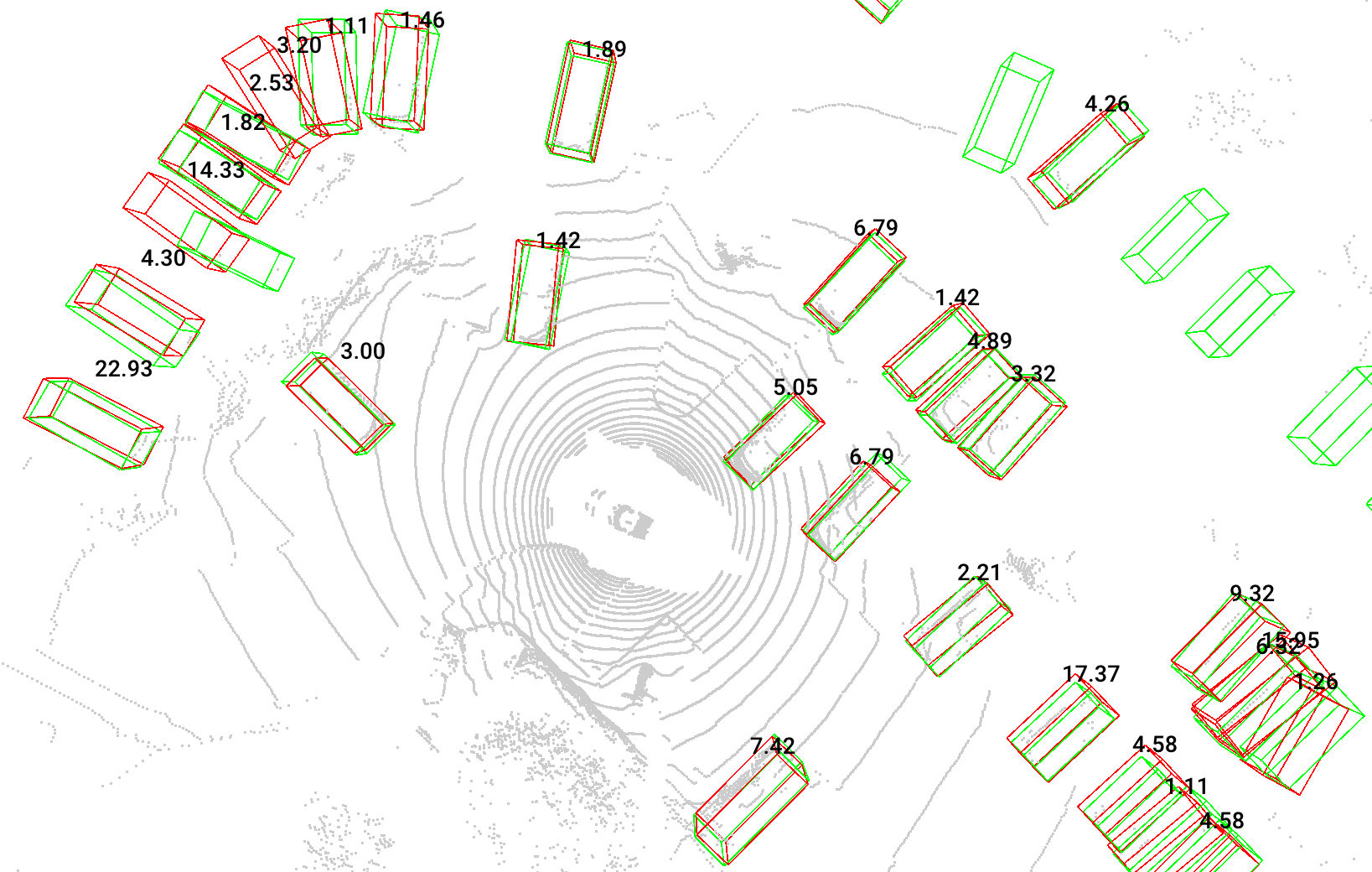}
	\caption{Visualization of uncertainty quantification of CenterPoint's predictions. Green bounding boxes are the ground truths, red bounding boxes are the predictions, and the numbers are the uncertainty values.}
	\label{lidar picture}
 \vspace{-5pt}
\end{figure}

\begin{table}[htbp]
\vspace{-5pt}
  \caption{Runtime comparison on CenterPoint.}
  \centering
  \begin{tabular}{l c c c}
  \toprule
    & MCI & GLM & MC-GLM \\
  \midrule
  Time(ms) & 525 & 10636 & 542\\
  \bottomrule
\end{tabular}
  \label{tab:running time}
\vspace{-5pt}
\end{table}

\subsection{Ablation Study}

All ablation studies are conducted on the CenterPoint detector. We analyze the sensitivity of MC-GLM to (i) offline Fisher-information approximations, including the approximation depth and the amount of training data used to estimate KFAC, and (ii) the online hyperparameters $\epsilon$ (finite-difference step) and $k$ (number of Monte-Carlo samples).

\textbf{(1) Fisher information depth.} In our setup, $H$, $N$ and $B$ denote head, neck and backbone
respectively. As shown in Tab.~\ref{tab:backbone neck head}, using Fisher information from deeper parts
of the network (up to $B{+}N{+}H$) improves the overall uncertainty--error alignment (higher $AvU$), but it
also increases memory consumption. Notably, while $AvU$ increases with depth, the gains in $P(A|C)$ are
small and $P(U|I)$ can slightly decrease, suggesting diminishing returns beyond the head.
In our CenterPoint experiments, the model size is 35MB, while the
KFAC Fisher storage is 173MB ($H$), 497MB ($N{+}H$) and 651MB ($B{+}N{+}H$).

\begin{table}[htbp]
\vspace{-5pt}
\caption{Ablation: Fisher information depth.}
\centering
\begin{tabular}{lcccc}
\toprule
                          &     & H   & N+H     & B+N+H \\ \hline
\multirow{2}{*}{$AvU$}    & IoU & 0.62 & 0.63        & 0.72         \\
                          & l2  & 0.62 & 0.63        & 0.73         \\ \hline
\multirow{2}{*}{$P(A|C)$} & IoU & 0.92 & 0.92        & 0.93         \\
                          & l2  & 0.95 & 0.95        & 0.95         \\ \hline
\multirow{2}{*}{$P(U|I)$} & IoU & 0.40 & 0.40        & 0.33         \\
                          & l2  & 0.40 & 0.42        & 0.34         \\
\bottomrule
\end{tabular}
\label{tab:backbone neck head}
\end{table}

\begin{table}[htbp]
\vspace{-5pt}
\caption{Ablation: training data proportion for KFAC.}
\centering
\begin{tabular}{lcccc}
\toprule
          & Dataset proportion   & 100\%       &  10\%       &  5\%  \\ \hline
\multirow{2}{*}{$AvU$}    & IoU  & 0.72         &  0.61       &  0.66  \\
                          & l2   & 0.73         &  0.62       &  0.67  \\ \hline
\multirow{2}{*}{$P(A|C)$} & IoU  & 0.93         &  0.92       &  0.92  \\
                          & l2   & 0.95         &  0.95       &  0.95  \\ \hline
\multirow{2}{*}{$P(U|I)$} & IoU  & 0.33         &  0.37       &  0.32  \\
                          & l2   & 0.34         &  0.39       &  0.35  \\
\bottomrule
\end{tabular}
\label{tab:different kfac}
\end{table}

\textbf{(2) Fisher information dataset size.} During the offline phase, the Fisher information matrix is typically estimated using the full training dataset. Since autonomous driving detectors are trained on large-scale data, this step can be expensive. We therefore estimate the KFAC matrix using a subset of the training data and evaluate the resulting MC-GLM uncertainty quality. As shown in \Cref{tab:different kfac}, using 10\% or 5\% of the training set yields comparable performance on $P(A|C)$ and $P(U|I)$, while $AvU$ drops more noticeably, indicating that Fisher estimation can be substantially subsampled with limited impact on the key uncertainty--error alignment metrics.

\textbf{(3) Finite-difference step $\epsilon$.} Tab.~\ref{tab:epsilon} compares different $\epsilon$ values. Considering all three metrics, the smallest step ($\epsilon=1\times 10^{-7}$) gives the best overall behavior, in particular substantially improving $P(U|I)$ while keeping $P(A|C)$ high. This aligns with the derivation in \eqref{fd_dirder}--\eqref{mcglm_est}: a smaller $\epsilon$ reduces the truncation error of the finite-difference approximation of the directional derivative $J(x)\eta$, making the MC-GLM covariance estimator closer to the GLM covariance $J(x)\Sigma_{\theta}J(x)^\top$.

\textbf{(4) Number of samples $k$.} Tab.~\ref{tab:sample times} reports performance for different $k$.
In the range $10$--$30$, results are similar, indicating that MC-GLM reaches a useful Monte Carlo accuracy with only a small number of posterior perturbations. This is consistent with Alg.~\ref{mcglm_pcode} and \eqref{mcglm_est}: the predictive covariance is estimated from the centered sample covariance of $k$ directional derivatives, and the resulting estimator is low-rank, so increasing $k$ beyond a modest value yields diminishing returns while directly increasing the number of forward passes. 
% We therefore use $k=10$ as the default trade-off between speed and stability.

\begin{table}[htbp]
\vspace{-5pt}
\caption{Ablation: finite-difference step $\epsilon$.}
\centering
\begin{tabular}{lcccc}
\toprule
                          & $\epsilon$ & $1.0$   & $1 \times 10^{-3}$  & $1 \times 10^{-7}$ \\ \hline
\multirow{2}{*}{$AvU$}    & IoU & 0.88 & 0.92        & 0.62         \\
                          & l2  & 0.90 & 0.94        & 0.62         \\ \hline
\multirow{2}{*}{$P(A|C)$} & IoU & 0.92 & 0.92        & 0.92         \\
                          & l2  & 0.94 & 0.94        & 0.95         \\ \hline
\multirow{2}{*}{$P(U|I)$} & IoU & 0.08 & 0.01        & 0.40         \\
                          & l2  & 0.08 & 0.01        & 0.40         \\
\bottomrule
\end{tabular}
\label{tab:epsilon}
\end{table}

\begin{table}[htbp]
\vspace{-5pt}
\caption{Ablation: number of Monte-Carlo samples $k$.}
\centering
\begin{tabular}{lcccc}
\toprule
                          & Sample times & 10  & 20  & 30 \\ \hline
\multirow{2}{*}{$AvU$}    & IoU & 0.62 & 0.64        & 0.63         \\
                          & l2  & 0.62 & 0.64        & 0.64         \\ \hline
\multirow{2}{*}{$P(A|C)$} & IoU & 0.92 & 0.92        & 0.92         \\
                          & l2  & 0.95 & 0.95        & 0.95         \\ \hline
\multirow{2}{*}{$P(U|I)$} & IoU & 0.40 & 0.39        & 0.38         \\
                          & l2  & 0.40 & 0.40        & 0.39         \\
\bottomrule
\end{tabular}
\label{tab:sample times}
% \vspace{-10pt}
\end{table}

\section{Conclusion}
We have introduced a new method MC-GLM to estimate instance-level predictive uncertainty for Laplace approximation based Bayesian neural networks. The currently available techniques of post hoc uncertainty estimation are unsuitable for instance-level uncertainty computation due to large run-time compute requirements, once calculation per instance. Our method fills this gap by estimating instance-level uncertainties using a preset number of forward passes. We conduct experiments with the bounding box uncertainty, which is a prime example of multiple instances and post hoc uncertainty prediction problem, to make our case.

\bibliographystyle{IEEEtran}
\bibliography{ijcnn}
\end{document}